\documentclass[conference]{IEEEtran}
\IEEEoverridecommandlockouts

\usepackage{cite}
\usepackage{amsmath,amssymb,amsfonts}
\usepackage{algorithmic}
\usepackage{graphicx}
\usepackage{textcomp}
\usepackage{xcolor}

\usepackage{booktabs}
\usepackage{multirow}
\usepackage{amsmath}
\usepackage[ruled,vlined]{algorithm2e}
\usepackage{tikz}

\usepackage{threeparttable}

\usepackage{pifont}
\usepackage{url}
\usepackage{enumitem}
\usepackage{makecell}
\usepackage{hyperref}

\def\BibTeX{{\rm B\kern-.05em{\sc i\kern-.025em b}\kern-.08em
    T\kern-.1667em\lower.7ex\hbox{E}\kern-.125emX}}
\begin{document}

\title{Understanding Cross-Sensor Feature Variations for Generalizable 3D Perception}

\author{
{\bf Xin Qiu}, {\bf Wenjie Liu}, {\bf Fuyuan Ai}, {\bf Yuchen Tan}, {\bf Zhiwei Xu}, {\bf Chunyi Song*}\thanks{$^{*}$Corresponding author.}\\
        Zhejiang University\\
        {qiuxinzju@zju.edu.cn}
}

\maketitle

\begin{abstract}
Radar-camera BEV perception often suffers from degraded performance when evaluated across datasets, as changes in driving scenes, sensor configurations, and environmental conditions can alter both the input observations and the internal fused representations. This work studies this issue from the perspective of source-domain variation modeling, aiming to improve the robustness of BEV-based 3D detectors without relying on target-domain samples. We introduce a framework that characterizes visual scene variations in the frequency domain and uses them to synthesize diverse source-domain views. By comparing the resulting fused BEV representations, the framework further captures how image-level variations influence multi-modal BEV features. These variation patterns are then used to regularize the detector, encouraging the learned fusion space to remain stable under latent scene changes. The proposed method is applied only during training and leaves the inference pipeline unchanged. Experiments on cross-dataset radar-camera 3D detection between View-of-Delft and TJ4DRadSet demonstrate consistent improvements over multiple BEV fusion backbones, and the gains remain effective when a small amount of target-domain data is available.

\end{abstract}


\section{Introduction}

Real-world multi-sensor perception data are inherently non-stationary, as changes in acquisition scenarios and environmental conditions induce shifts in both sensor observations and learned feature distributions~\cite{feng2020deep,bijelic2020seeing,sun2022shift,li2024unimode}. These shifts severely hinder transfer from source datasets to unseen target domains, making it crucial to mine latent sensor feature shifts from source data and learn domain-generalizable representations~\cite{li2023bev,li2022unsupervised,jiang2024bev}. Autonomous driving perception is a representative case, where data from different cities, road structures, weather conditions, sensor configurations, and platforms exhibit substantial cross-scenario discrepancies~\cite{zheng2023cross,wenzel20204seasons,li2024domain}. Compared with single-sensor tasks, multi-sensor perception must further handle modality-specific statistics, spatial correspondence, and fusion-induced feature shifts, making cross-scenario generalization more challenging.

This challenge becomes more concrete in radar-camera BEV 3D detection~\cite{ICLR2025_21dabaac,stacker2023rc,lin2024rcbevdet}. Camera images provide semantic and texture cues, while radar point clouds offer physical measurements such as range, velocity, and geometry. After encoding, projection, and fusion, these modalities form BEV representations for 3D object detection. However, cross-dataset scenario variations first alter image-level visual statistics, including illumination, texture, contrast, and high-frequency details, and these changes are propagated through image encoding, BEV projection, and radar-camera fusion~\cite{jiang2024bev,liu2024bevuda,zhang2024bevuda++}. Therefore, cross-dataset generalization is not merely an input-level style transfer problem, but a propagation modeling problem from sensor observation shifts to fused BEV feature shifts.

Existing domain generalization methods improve robustness through data augmentation, domain-invariant constraints, or architectural design~\cite{mai2025domain,muandet2013domain,wang2022generalizing,shui2022benefits}. However, for radar-camera BEV detection, addressing domain shifts only at the input or final feature level is insufficient: the former does not reveal whether visual variations affect fused BEV representations, while the latter ignores the source of the shift~\cite{lu2025towards,jiang2024bev,liu2024bevuda}. Cross-dataset generalization requires modeling how sensor observation shifts propagate into BEV feature shifts. Based on the above analysis, we studies cross-dataset domain generalization from the perspective of sensor-to-feature shift mining. Cross-scenario variations propagate through image encoding, BEV projection, and cross-modal fusion, eventually manifesting as feature shifts in the fused BEV space. Since image spectral statistics capture visual-side differences such as illumination, texture, contrast, and imaging style, we mine latent visual shift patterns from source domains and model their propagation to BEV features, thereby constructing more cross-scenario representative training constraints to improve generalization to unseen datasets without accessing target-domain data.

To this end, we propose a Visual-to-BEV Scene Shift Mining (\textbf{VBS$^2$M}) framework. First, we mine scene-level spectral prototypes from whole-image frequency statistics in source-domain images and generate scene-shifted views to simulate latent visual scenario variations. Then, the original and scene-shifted images are separately paired with the same radar input and fed into the BEV detector, where their fused BEV representations are compared to explicitly model the propagation from visual scene shifts to BEV feature shifts. Furthermore, we mine BEV scene shift prototypes from BEV shift descriptors and introduce prototype-guided BEV scene regularization, encouraging the detector to learn fusion representations that are more stable under cross-dataset scenario changes. To the best of our knowledge, VBS$^2$M is the first framework to study domain generalization for radar-camera BEV 3D detection from the perspective of visual-to-BEV scene shift mining.

Extensive experiments on bidirectional transfer between View-of-Delft~\cite{palffy2022multi} and TJ4DRadSet~\cite{zheng2022tj4dradset} demonstrate the effectiveness of VBS$^2$M. Across BEVFusion~\cite{liu2023bevfusion}, RaCFormer~\cite{chu2025racformer}, and RCBEVDet~\cite{lin2024rcbevdet}, our method improves OOD detection performance over source-only training and domain generalization baselines. It also provides stable gains under few-shot target-domain settings with 10\%, 20\%, and 30\% labeled target data. Ablation studies and visualization analyses further show that the improvements are driven by data-driven spectral scene mining, BEV shift prototype modeling, and more stable visual-to-BEV feature propagation.

The main contributions are summarized as follows:
\begin{itemize}
    \item First, we recast cross-dataset radar-camera BEV detection as a sensor-to-feature shift mining problem.

    \item Second, we mine scene-level spectral prototypes from source-domain image frequency statistics to generate data-driven scene-shifted views.

    \item Third, we discover BEV scene shift prototypes to model how visual shifts propagate through radar-camera fusion.

    \item Fourth, we introduce prototype-guided BEV regularization to learn scenario-stable fusion representations without target-domain data.
\end{itemize}

\section{Related Work}
\subsection{Domain Generalization under Distribution Shift}
Domain generalization aims to train models using only source-domain data while maintaining stable performance on target domains~\cite{wang2022generalizing,zhang2024bevuda++,muandet2013domain,huang2025bridging}. Unlike domain adaptation, it cannot access target-domain data during training, requiring models to learn transferable representations from source domains~\cite{liu2024bevuda,shu2021open}. Existing studies mitigate distribution shifts through data augmentation, domain-invariant representation learning, feature regularization, and risk balancing, achieving progress in image classification, semantic segmentation, and object detection~\cite{schwonberg2023augmentation,lee2024object,krueger2021out,he2025boosting,li2026towards}. However, most methods focus on unimodal visual data and model distribution shift as either input appearance variation or high-level feature statistic changes. For multi-sensor perception, cross-dataset discrepancies arise not only from individual sensor observations, but also propagate through cross-modal encoding, spatial projection, and feature fusion, leading to shifts in the fused representation space. Therefore, constraining only the input space or final feature space is insufficient to characterize the propagation mechanism of multi-sensor domain shifts, limiting generalization in cross-dataset scenarios~\cite{li2026dataset,feng2026towards}.

\subsection{Autonomous Driving Radar-Camera BEV Perception}
Radar-camera BEV perception is an important direction for 3D object detection in autonomous driving~\cite{qiu2026rpgfusion,zhong2025cvfusion,li2025rctrans,xia2026r4det,li2026sdef}. Cameras provide dense semantic, texture, and appearance cues, while radar offers range, velocity, and geometric measurements, showing advantages under low-light and adverse weather conditions. By projecting and fusing both modalities into a unified BEV space, models can jointly exploit visual semantics and radar geometry for more robust object localization and recognition~\cite{chu2025racformer,li2026sdef,zhong2025cvfusion,stacker2023rc,ICLR2025_21dabaac}.
However, radar-camera BEV perception suffers from data scarcity~\cite{sheeny2021radiate,wu2024survey}. High-quality data collection requires synchronized and calibrated multi-sensor platforms, while 3D annotation demands accurate spatial boxes and cross-sensor alignment, making dataset construction costly. Existing public datasets remain limited in scale, scene diversity, and sensor configurations, and thus cannot fully cover cross-scenario variations in real deployment~\cite{zheng2022tj4dradset,palffy2022multi}. Under this condition, improving cross-dataset generalization is crucial: models must not only fuse radar and camera information effectively, but also learn BEV representations that remain stable under scenario changes from limited source-domain data.

\section{Preliminary}
\subsection{Radar-Camera BEV Detection System}
Radar-camera BEV 3D detection takes camera images $I$ and radar point clouds $P$ as inputs, and performs multi-modal fusion and object detection in a unified bird's-eye-view (BEV) space. Given the input pair $(I, P)$, the model first extracts visual and radar features using an image encoder and a radar encoder:
$
F_I = E_I(I),  F_R = E_R(P),
$
where $F_I$ contains semantic, texture, and appearance information from images, while $F_R$ encodes radar measurements such as spatial location, range, velocity, and reflection intensity.

Since cameras and radar sensors operate in different observation spaces, their features need to be transformed into a unified BEV space~\cite{jiang2024bev,lin2024rcbevdet,zhang2024bevuda++}. Image features are usually mapped to the BEV plane through depth estimation, view transformation, or query-based projection, while radar features can be converted into BEV representations via voxelization, pillar encoding, or point scattering~\cite{liu2023bevfusion,lin2024rcbevdet,chu2025racformer}.

This process can be formulated as:
$
B_I = T_I(F_I),  B_R = T_R(F_R),
$
where $T_I(\cdot)$ and $T_R(\cdot)$ denote the image-to-BEV and radar-to-BEV transformation modules, respectively.
The model then integrates image BEV features and radar BEV features into a unified multi-modal BEV representation:
\begin{equation}
f_B = \mathcal{F}(B_I, B_R), \qquad \hat{y} = h_{\theta}(f_B),
\end{equation}
where $\mathcal{F}(\cdot)$ denotes the cross-modal fusion function, and $h_{\theta}(\cdot)$ is the detection head.
The prediction $\hat{y}$ includes object categories, 3D locations, sizes, orientations, and velocities.
Therefore, the overall radar-camera BEV detection pipeline can be summarized as:
\begin{equation}
\hat{y}
=
h_{\theta}
\left(
\mathcal{F}
\left(
T_I(E_I(I)),
T_R(E_R(P))
\right)
\right).
\end{equation}
This formulation shows that the final detection results depend not only on camera and radar observations at the input level, but also on image encoding, radar encoding, BEV projection, and cross-modal fusion.

\subsection{Domain Generalization for 3D Object Detection}
Domain generalization aims to train a model using only source-domain data such that it can maintain stable performance on unseen target domains~\cite{he2025boosting,li2026towards,krueger2021out,schwonberg2023augmentation}. Let $d \in \mathcal{D}$ denote a data domain, where different domains correspond to different data-generating distributions. For radar-camera BEV detection, each sample consists of a camera image, a radar point cloud, and 3D detection annotations, denoted as $(I, P, y) \sim p_d(I, P, y)$. Given a detection model $F_{\theta}$, its expected risk on domain $d$ is defined as:
\begin{equation}
R_d(\theta)
=
\mathbb{E}_{(I,P,y)\sim p_d}
\left[
\ell\left(F_{\theta}(I,P), y\right)
\right],
\end{equation}
where $\ell(\cdot)$ denotes the detection loss.
Assume that the training data are drawn from a set of source domains $\mathcal{D}_s$, while the target domains belong to an unseen domain set $\mathcal{D}_t$. The ideal objective of domain generalization is to minimize the expected risk on unseen target domains:
$
\min_{\theta}
\ \mathbb{E}_{d\sim \mathcal{D}_t}
\left[
R_d(\theta)
\right].
$

However, target-domain data are inaccessible during training, making the target risk impossible to optimize directly. In practice, the model can only learn transferable representations from source-domain data by optimizing the source risk, with the goal of improving target-domain generalization:
\begin{equation}
\theta^{*}
=
\arg\min_{\theta}
\mathbb{E}_{d\sim \mathcal{D}_s}
\left[
R_d(\theta)
\right],
\mathcal{G}(\theta^{*})
=
\mathbb{E}_{d\sim \mathcal{D}_t}
\left[
R_d(\theta^{*})
\right],
\end{equation}

In cross-dataset radar-camera BEV detection, the source and target domains usually satisfy:
\begin{equation}
p_s(I,P,y) \neq p_t(I,P,y),
f_B = g_{\theta}(I,P),
p_s(f_B) \neq p_t(f_B),
\end{equation}
where $g_{\theta}(\cdot)$ denotes the feature extraction and fusion process that maps multi-sensor inputs to the fused BEV representation $f_B$. Therefore, the key challenge of cross-dataset domain generalization is to learn BEV representations that remain stable under potential scenario shifts from source domains alone, thereby reducing the risk on unseen target domains.

\section{Method}
\begin{algorithm}[t]
\caption{VBS$^2$M: Visual-to-BEV Sensor Shift Mining}
\label{alg:vbs2m}
\KwIn{Source dataset $\mathcal{D}_s=\{(I_i,P_i,y_i)\}_{i=1}^{N}$, image prototypes $C^I$, BEV prototypes $C^B$, detector $h_\theta\circ g_\theta$}
\KwOut{Trained detector $h_\theta\circ g_\theta$}

Initialize detector parameters $\theta$ and prototypes $C^I,C^B$\;

\ForEach{training step}{
    Sample a minibatch $\{(I,P,y)\}$ from $\mathcal{D}_s$\;
    
    Extract image spectral descriptor $z_I$ from the amplitude spectrum of $I$\;
    
    Assign $z_I$ to image scene prototypes $C^I$ and generate scene-shifted image $\hat{I}$\;
    
    Compute BEV features:
    $f_B=g_\theta(I,P)$ and $\hat{f}_B=g_\theta(\hat{I},P)$\;
    
    Construct BEV shift descriptor $z_B=\phi(\hat{f}_B-f_B,f_B,\hat{f}_B)$\;
    
    Assign $z_B$ to BEV scene prototypes $C^B$ and obtain BEV shift direction $d_B$\;
    
    Compute sensitivity $\rho=1-\mathrm{Cos}(f_B,\hat{f}_B)$\;
    
    Construct prototype-guided BEV feature:
    $\tilde{f}_B=f_B+\lambda_B\rho\cdot\mathrm{Broadcast}(d_B)$\;
    
    Predict $\hat{y}=h_\theta(\tilde{f}_B)$\;
    
    Update $\theta$ with standard detection loss:
    $\mathcal{L}=\mathcal{L}_{det}(\hat{y},y)$\;
}

\Return{$h_\theta\circ g_\theta$}\;
\end{algorithm}
\subsection{Image Spectral Scene Prototype Mining}

Cross-dataset scenario variations often induce changes in image statistics~\cite{zheng2023cross,wenzel20204seasons,sun2022shift}. To capture global scene information, we operate in the frequency domain. Given an input image $I$, we compute its 2D Fourier transform:
$
F(I) = A(I)e^{j\Phi(I)},
$
where $A(I)$ and $\Phi(I)$ denote the amplitude and phase spectra, respectively.
From the log-amplitude spectrum, we extract a spectral descriptor:
\begin{equation}
z_I = \psi(\log(A(I)+\epsilon)),
\end{equation}
where $\psi(\cdot)$ is a spectral pooling function that compresses statistics over the frequency plane.
We maintain a set of $K$ learnable spectral scene prototypes:
$
C_I = \{c_1^I, c_2^I, \dots, c_K^I\}.
$
The soft assignment of the descriptor $z_I$ to each prototype is:
\begin{equation}
q_k^I = \frac{\exp(\mathrm{sim}(z_I, c_k^I)/\tau_I)}{\sum_{j=1}^K \exp(\mathrm{sim}(z_I, c_j^I)/\tau_I)},
\end{equation}
where $\mathrm{sim}(\cdot,\cdot)$ denotes cosine similarity and $\tau_I$ is a temperature.
Based on these assignments, the spectral modulation mask is computed as:
\begin{equation}
M_I = \sum_{k=1}^K q_k^I \cdot \mathrm{Reshape}(W_I c_k^I),
\end{equation}
with $W_I$ a lightweight mapping network.
The amplitude spectrum is modulated via:
\begin{equation}
\hat{A}(I) = A(I) \odot \exp(\lambda_I M_I),
\end{equation}
where $\lambda_I$ controls the strength of scene variation.
Finally, the scene-shifted image is reconstructed while keeping the phase spectrum unchanged:
$
\hat{I} = \mathcal{F}^{-1}(\hat{A}(I)e^{j\Phi(I)}).
$

The resulting $\hat{I}$ provides a source-domain image under a latent visual scene, effectively expanding the source-domain scene distribution.

\begin{figure*}[t]
    \centering
    \includegraphics[width=\textwidth]{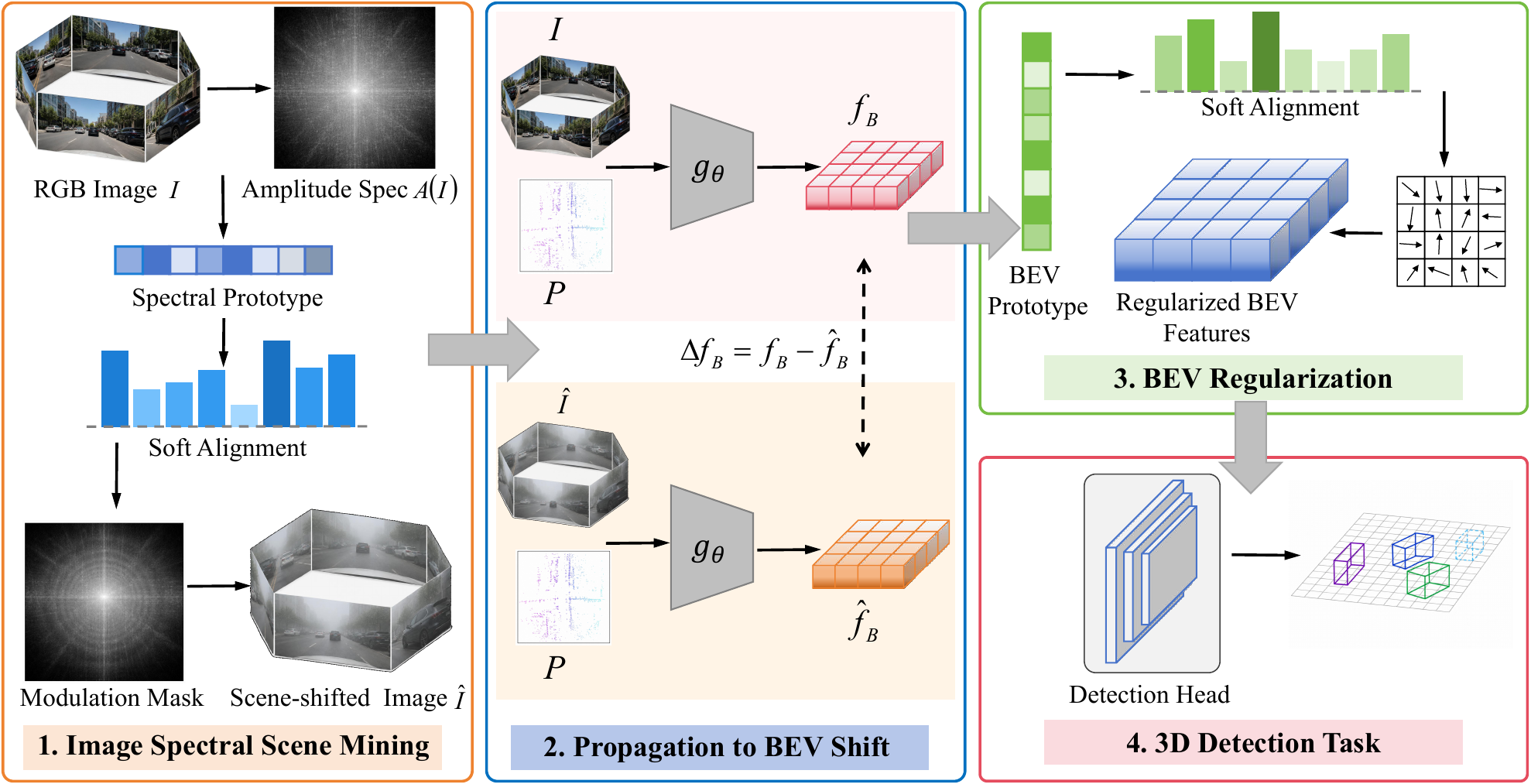}
    \caption{
Overview of VBS$^2$M.
The framework includes image spectral scene mining, propagation to BEV shift, BEV regularization, and the final 3D detection task.
It mines spectral scene prototypes from source-domain images, models how visual shifts propagate into fused BEV features, and regularizes BEV representations to improve cross-dataset radar-camera BEV detection.
}
    \label{fig:main}
\end{figure*}

\subsection{Propagation from Image Scene Shifts to BEV Shifts}

Generating scene-shifted views only at the image level is insufficient to characterize cross-domain shifts in radar-camera BEV detection~\cite{li2026dataset}. Since the final predictions of a BEV detector are determined by the fused BEV representation, it is necessary to further analyze how image-level scene variations affect the BEV representation space after image encoding, BEV projection, and cross-modal fusion.

Given the original image $I$, the scene-shifted image $\hat{I}$, and the radar point cloud $P$, we feed the original and scene-shifted inputs into the same radar-camera BEV detector:
\begin{equation}
f_B = g_{\theta}(I, P),\quad
\hat{f}_B = g_{\theta}(\hat{I}, P),
\end{equation}
where the radar point cloud $P$ is kept unchanged. Here, $f_B$ denotes the fused BEV representation obtained from the original image-radar pair, while $\hat{f}_B$ denotes the fused BEV representation after image scene variation.

The difference between the two representations indicates how the image scene shift propagates into the BEV space:
$
\Delta f_B = \hat{f}_B - f_B .
$
To model this propagated shift, we compress the high-dimensional BEV difference into a BEV scene shift descriptor:
$
z_B = \phi(\Delta f_B, f_B, \hat{f}_B),
$
where $\phi(\cdot)$ is a lightweight statistical mapping function. In this work, we adopt a simple yet effective implementation:
\begin{equation}
z_B =
\left[
\mathrm{GAP}(\Delta f_B),
\mathrm{GAP}(|\Delta f_B|),
1-\mathrm{Cos}(f_B,\hat{f}_B)
\right],
\end{equation}
where $\mathrm{GAP}(\Delta f_B)$ describes the average direction of the BEV representation shift, $\mathrm{GAP}(|\Delta f_B|)$ measures its overall magnitude, and $1-\mathrm{Cos}(f_B,\hat{f}_B)$ quantifies the global discrepancy between the original and scene-shifted BEV representations.

\subsection{Prototype-Guided BEV Scene Regularization}

To encourage the detector to learn stable fused BEV representations, we construct a scene-regularized BEV feature based on the mined BEV scene shift directions. Given the original BEV feature $f_B$, the BEV scene shift direction $d_B$, and the current sample's sensitivity to image-level scene variation, the prototype-guided BEV feature is formulated as:
\begin{equation}
\tilde{f}_B = f_B + \lambda_B \, \rho \cdot \mathrm{Broadcast}(d_B),
\end{equation}
where $\lambda_B$ controls the intensity of the BEV scene shift, and $\mathrm{Broadcast}(\cdot)$ expands $d_B$ to match the spatial dimensions of the BEV feature. The coefficient $\rho$ measures the sensitivity of the BEV representation to image-level scene variation:
\begin{equation}
\rho = 1 - \mathrm{Cos}(f_B, \hat{f}_B),
\end{equation}
where $\hat{f}_B$ denotes the BEV feature extracted from the image scene-shifted input. A larger $\rho$ indicates that the current sample is more sensitive to visual scene changes, requiring stronger BEV scene regularization.
During training, the detector uses a single BEV feature for prediction. Specifically, the BEV feature is selected as:
\begin{equation}
f_B^{\text{train}} =
\begin{cases}
f_B, & \text{without scene regularization}, \\
\tilde{f}_B, & \text{with prototype-guided scene regularization}.
\end{cases}
\end{equation}
The final detection output is then obtained by
$
\hat{y} = h_{\theta}(f_B^{\text{train}}).
$
In this way, the proposed module regularizes the BEV representation during training while keeping the optimization objective identical to the standard detection loss. 

\subsection{Training and Inference}

During training, the model is optimized with the standard detection loss:
$
\mathcal{L}=\ell(\hat{y},y),
$
where \(\hat{y}=h_{\theta}(f_B^{train})\). The proposed spectral scene mining and prototype-guided BEV regularization are only used to construct \(f_B^{train}\) during training.

During inference, we directly use the original BEV feature \(f_B=g_{\theta}(I,P)\) and predict \(\hat{y}=h_{\theta}(f_B)\). No spectral modulation, BEV shift mining, or prototype-guided regularization is performed, so the inference pipeline remains unchanged.
\section{Experiment}
\subsection{Experimental Setup}
\paragraph{Models}
We consider three representative radar-camera BEV fusion models, including RCBEVDet~\cite{lin2024rcbevdet}, BEVFusion~\cite{liu2023bevfusion}, and RaCFormer~\cite{chu2025racformer}, all of which have demonstrated strong and competitive performance in 3D object detection. Our method is designed as a plug-and-play training framework and can be directly integrated into these detectors without modifying their backbone architectures, detection heads, or inference pipelines.

\paragraph{Baselines}
We compare our method with both domain generalization baselines and radar-camera fusion baselines. The domain generalization methods include DG-GCD~\cite{rathore2025domain}, CPerb~\cite{zhao2024novel}, SPG~\cite{xu2021spg}, and VL2V-ADiP~\cite{addepalli2024leveraging}, which improve out-of-domain performance. However, most of these methods are designed from a general visual domain generalization perspective, mainly focusing on distribution alignment in the input space or high-level feature space. They lack modeling of how sensor observation shifts propagate into fused BEV representation shifts in radar-camera BEV detection.

\begin{table*}[t]
\centering
\caption{Cross-dataset generalization results of \textbf{BEVFusion} in terms of mAP (\%). Oracle denotes training and evaluation on the same target dataset and serves as the \textbf{upper-bound }reference.}
\label{tab:cross_dataset_bevfusion}
\resizebox{\textwidth}{!}{
\begin{tabular}{l|cccc|cccc|cccc|cccc}
\toprule
\multirow{3}{*}{\textbf{BEVFusion}}
& \multicolumn{8}{c|}{\textbf{VoD $\rightarrow$ TJ4DRadSet}}
& \multicolumn{8}{c}{\textbf{TJ4DRadSet $\rightarrow$ VoD}} \\
\cmidrule(lr){2-9} \cmidrule(lr){10-17}
& \multicolumn{4}{c|}{\textbf{VoD (Source)}}
& \multicolumn{4}{c|}{\textbf{TJ4DRadSet (Target)}}
& \multicolumn{4}{c|}{\textbf{TJ4DRadSet (Source)}}
& \multicolumn{4}{c}{\textbf{VoD (Target)}} \\
& Car & Ped & Cyc & mAP
& Car & Ped & Cyc & mAP
& Car & Ped & Cyc & mAP
& Car & Ped & Cyc & mAP \\
\midrule
Oracle
& - & - & - & -
& 27.51 & 25.48 & 53.61 & 35.53
& - & - & - & -
& 37.85 & 40.96 & 68.95 & 49.25 \\

Source Only
& 37.85 & 40.96 & 68.95 & 49.25
& 16.52 & 14.37 & 35.18 & 22.02
& 27.51 & 25.48 & 53.61 & 35.53
& 23.63 & 28.15 & 44.36 & 32.05 \\

DG-GCD
& 36.91 & 40.52 & 66.84 & 48.09
& 15.94 & 14.82 & 35.73 & 22.16
& 26.83 & 24.37 & 52.18 & 34.46
& 24.12 & 28.34 & 45.07 & 32.51 \\

CPerb
& 37.24 & 40.71 & 67.32 & 48.42
& 16.08 & 14.95 & 35.96 & 22.33
& 26.95 & 24.61 & 52.46 & 34.67
& 24.45 & 28.57 & 45.39 & 32.80 \\

SPG
& 37.46 & 40.88 & 67.71 & 48.68
& 16.21 & 15.12 & 36.14 & 22.49
& 27.12 & 24.84 & 52.73 & 34.90
& 24.76 & 28.73 & 45.72 & 33.07 \\

VL2V-ADiP
& 36.52 & 41.15 & 65.39 & 47.69
& 16.35 & 15.47 & 36.50 & 22.77
& 27.05 & 23.88 & 52.47 & 34.47
& 25.11 & 28.92 & 46.18 & 33.40 \\

VBS$^2$M (Ours)
& \textbf{38.32} & \textbf{42.18} & \textbf{70.41} & \textbf{50.30}
& \textbf{22.36} & \textbf{19.21} & \textbf{40.57} & \textbf{27.38}
& \textbf{27.62} & \textbf{26.37} & \textbf{54.26} & \textbf{36.08}
& \textbf{30.12} & \textbf{31.86} & \textbf{51.04} & \textbf{37.67} \\
\bottomrule
\end{tabular}}
\end{table*}

\begin{table*}[t]
\centering
\caption{Cross-dataset generalization results of \textbf{RaCFormer} in terms of mAP (\%). Oracle denotes training and evaluation on the same target dataset and serves as the \textbf{upper-bound }reference.}
\label{tab:cross_dataset_racformer}
\resizebox{\textwidth}{!}{
\begin{tabular}{l|cccc|cccc|cccc|cccc}
\toprule
\multirow{3}{*}{\textbf{RaCFormer}}
& \multicolumn{8}{c|}{\textbf{VoD $\rightarrow$ TJ4DRadSet}}
& \multicolumn{8}{c}{\textbf{TJ4DRadSet $\rightarrow$ VoD}} \\
\cmidrule(lr){2-9} \cmidrule(lr){10-17}
& \multicolumn{4}{c|}{\textbf{VoD (Source)}}
& \multicolumn{4}{c|}{\textbf{TJ4DRadSet (Target)}}
& \multicolumn{4}{c|}{\textbf{TJ4DRadSet (Source)}}
& \multicolumn{4}{c}{\textbf{VoD (Target)}} \\
& Car & Ped & Cyc & mAP
& Car & Ped & Cyc & mAP
& Car & Ped & Cyc & mAP
& Car & Ped & Cyc & mAP \\
\midrule
Oracle
& - & - & - & -
& 53.36 & 27.08 & 40.56 & 40.33
& - & - & - & -
& 47.30 & 46.21 & 69.80 & 54.44 \\

Source Only
& 47.30 & 46.21 & 69.80 & 54.44
& 34.28 & 15.70 & 25.16 & 25.05
& 53.36 & 27.08 & 40.56 & 40.33
& 29.31 & 31.15 & 37.48 & 32.65 \\

DG-GCD
& 46.74 & 44.86 & 69.12 & 53.57
& 33.72 & 15.38 & 25.41 & 24.84
& 51.92 & 26.64 & 39.75 & 39.44
& 27.85 & 29.84 & 38.16 & 31.95 \\

CPerb
& 46.95 & 45.18 & 69.43 & 53.85
& 33.96 & 15.52 & 25.57 & 25.02
& 52.16 & 26.89 & 39.92 & 39.66
& 28.42 & 30.26 & 38.37 & 32.35 \\

SPG
& 47.08 & 45.63 & 69.68 & 54.13
& 34.07 & 15.75 & 25.69 & 25.17
& 52.53 & 27.12 & 40.08 & 39.91
& 28.96 & 30.73 & 38.61 & 32.77 \\

VL2V-ADiP
& 47.14 & 44.23 & 70.10 & 53.82
& 34.19 & 15.96 & 25.83 & 25.33
& 51.47 & 27.58 & 39.32 & 39.46
& 26.50 & 29.53 & 38.95 & 31.66 \\

VBS$^2$M (Ours)
& \textbf{48.46} & \textbf{46.81} & \textbf{71.22} & \textbf{55.50}
& \textbf{39.04} & \textbf{18.71} & \textbf{31.36} & \textbf{29.70}
& \textbf{54.12} & \textbf{28.67} & \textbf{41.42} & \textbf{41.40}
& \textbf{36.08} & \textbf{35.02} & \textbf{43.47} & \textbf{38.19} \\
\bottomrule
\end{tabular}}
\end{table*}

\begin{table*}[t]
\centering
\caption{Cross-dataset generalization results of \textbf{RCBEVDet} in terms of mAP (\%). Oracle denotes training and evaluation on the same target dataset and serves as the \textbf{upper-bound }reference.}
\label{tab:cross_dataset_rcbevdet}
\resizebox{\textwidth}{!}{
\begin{tabular}{l|cccc|cccc|cccc|cccc}
\toprule
\multirow{3}{*}{\textbf{RCBEVDet}}
& \multicolumn{8}{c|}{\textbf{VoD $\rightarrow$ TJ4DRadSet}}
& \multicolumn{8}{c}{\textbf{TJ4DRadSet $\rightarrow$ VoD}} \\
\cmidrule(lr){2-9} \cmidrule(lr){10-17}
& \multicolumn{4}{c|}{\textbf{VoD (Source)}}
& \multicolumn{4}{c|}{\textbf{TJ4DRadSet (Target)}}
& \multicolumn{4}{c|}{\textbf{TJ4DRadSet (Source)}}
& \multicolumn{4}{c}{\textbf{VoD (Target)}} \\
& Car & Ped & Cyc & mAP
& Car & Ped & Cyc & mAP
& Car & Ped & Cyc & mAP
& Car & Ped & Cyc & mAP \\
\midrule
Oracle
& - & - & - & -
& 48.72 & 24.63 & 38.91 & 37.42
& - & - & - & -
& 40.63 & 38.86 & 70.48 & 49.99 \\

Source Only
& 40.63 & 38.86 & 70.48 & 49.99
& 29.84 & 13.92 & 24.36 & 22.71
& 48.72 & 24.63 & 38.91 & 37.42
& 24.56 & 26.84 & 43.27 & 31.56 \\

DG-GCD
& 39.72 & 37.94 & 68.76 & 48.81
& 29.37 & 14.18 & 24.71 & 22.75
& 47.31 & 23.82 & 37.64 & 36.26
& 25.02 & 27.16 & 44.05 & 32.08 \\

CPerb
& 40.05 & 38.21 & 69.34 & 49.20
& 29.55 & 14.37 & 24.96 & 22.96
& 47.68 & 24.05 & 38.02 & 36.58
& 25.38 & 27.42 & 44.38 & 32.39 \\

SPG
& 40.28 & 38.47 & 69.72 & 49.49
& 29.76 & 14.58 & 25.18 & 23.17
& 48.05 & 24.31 & 38.37 & 36.91
& 25.77 & 27.81 & 44.72 & 32.77 \\

VL2V-ADiP
& 39.86 & 38.55 & 69.91 & 49.44
& 30.12 & 14.86 & 25.74 & 23.57
& 47.92 & 24.18 & 38.25 & 36.78
& 26.31 & 28.24 & 45.36 & 33.30 \\

VBS$^2$M (Ours)
& \textbf{41.12} & \textbf{39.35} & \textbf{71.26} & \textbf{50.58}
& \textbf{34.08} & \textbf{17.42} & \textbf{29.86} & \textbf{27.12}
& \textbf{49.35} & \textbf{25.42} & \textbf{40.17} & \textbf{38.31}
& \textbf{30.14} & \textbf{31.53} & \textbf{49.72} & \textbf{37.13} \\
\bottomrule
\end{tabular}}
\end{table*}

\paragraph{Evaluation Setup}
To evaluate cross-dataset generalization, we conduct bidirectional transfer experiments on View-of-Delft (VoD)~\cite{palffy2022multi} and TJ4DRadSet~\cite{zheng2022tj4dradset}. Specifically, we consider two transfer directions, VoD $\rightarrow$ TJ4DRadSet and TJ4DRadSet $\rightarrow$ VoD, where the model is trained on one dataset and directly tested on the other. This setting reflects the model's transferability across different acquisition platforms, road scenarios, object distributions, and sensor configurations. Since both datasets contain radar-camera perception data and share similar 3D detection categories, we evaluate on the common classes, including Car, Pedestrian, and Cyclist, and report class-wise AP and overall mAP.

We also evaluate cross-dataset adaptation under different target-domain  ratios. When limited labeled target-domain data are available, we use 10\%, 20\%, and 30\% of the target training set together with the full source training set for training, and then evaluate the model on the full target test set. 

\paragraph{Implementation Details}
For all baseline detectors, we keep the original network architecture, detection head, and inference pipeline unchanged, and introduce VBS$^2$M only during training. The number of image spectral prototypes is set to $K=8$, the temperature is set to $\tau_I=0.2$, and the spectral modulation strength is set to $\lambda_I=0.1$, with a probability of $0.5$ for each mini-batch. For BEV scene shift modeling, the number of BEV prototypes is set to $L=8$, the temperature is set to $\tau_B=0.2$, and the BEV regularization strength is set to $\lambda_B=0.05$. During inference, image spectral modulation, prototype assignment, and BEV scene regularization are disabled, so the inference pipeline remains identical to that of the base detector.

\subsection{Zero-Shot Target-Domain Results}
In this setting, the model is trained only on the source dataset and directly evaluated on the unseen target dataset without using any target-domain samples during training.

From the source-domain results, VBS$^2$M consistently improves the performance of all three BEV fusion detectors, indicating that our method enhances cross-scenario generalization without degrading the detection ability on the source dataset. As shown in Tables~\ref{tab:cross_dataset_bevfusion}, \ref{tab:cross_dataset_racformer}, and \ref{tab:cross_dataset_rcbevdet}, when trained on VoD and evaluated on the VoD source domain, VBS$^2$M improves the mAP of BEVFusion, RaCFormer, and RCBEVDet to 50.30\%, 55.50\%, and 50.58\%, respectively, outperforming Source Only and other domain generalization baselines. Similarly, on the TJ4DRadSet source domain, VBS$^2$M also achieves the highest source-domain mAP, reaching 36.08\%, 41.40\%, and 38.31\% for the three detectors. These results suggest that mining visual scene shifts and BEV representation shifts within the source domain does not introduce destructive perturbations. Instead, it provides structured training constraints that help the model learn more stable and effective fused BEV representations.

The improvements are more pronounced on unseen target domains, demonstrating that VBS$^2$M effectively alleviates BEV representation shifts caused by cross-scenario discrepancies. In the VoD $\rightarrow$ TJ4DRadSet direction, VBS$^2$M improves the target-domain mAP of BEVFusion, RaCFormer, and RCBEVDet to 27.38\%, 29.70\%, and 27.12\%, respectively, yielding gains of 5.36, 4.65, and 4.41 percentage points over Source Only. It also clearly outperforms the strongest domain generalization baseline. In the reverse TJ4DRadSet $\rightarrow$ VoD direction, VBS$^2$M again achieves the best results, with mAP values of 37.67\%, 38.19\%, and 37.13\%, corresponding to improvements of 5.62, 5.54, and 5.57 percentage points over Source Only. The gains in both transfer directions indicate that cross-dataset performance degradation is not only caused by input-level distribution shifts, but also by the propagation of image scene shifts into the fused BEV representation space. By modeling this process through visual-to-BEV scene shift mining, our method achieves robust generalization improvements across different detectors and transfer directions.

\begin{table*}[t]
\centering
\caption{Few-shot target-domain generalization on VoD $\rightarrow$ TJ4DRadSet in terms of AP/mAP (\%). 
The model is trained on the full VoD source dataset and additionally uses different proportions of labeled TJ4DRadSet target-domain training data.}
\label{tab:fewshot_vod_to_tj}
\resizebox{\textwidth}{!}{
\renewcommand{\arraystretch}{1.05}
\begin{tabular}{ll|cccc|cccc|cccc|cccc}
\toprule
\multirow{2}{*}{\textbf{Model}} 
& \multirow{2}{*}{\textbf{Setting}}
& \multicolumn{4}{c|}{\textbf{0\% Target}}
& \multicolumn{4}{c|}{\textbf{10\% Target}}
& \multicolumn{4}{c|}{\textbf{20\% Target}}
& \multicolumn{4}{c}{\textbf{30\% Target}} \\
\cmidrule(lr){3-6} \cmidrule(lr){7-10} \cmidrule(lr){11-14} \cmidrule(lr){15-18}
& 
& \textbf{Car} & \textbf{Ped} & \textbf{Cyc} & \textbf{mAP}
& \textbf{Car} & \textbf{Ped} & \textbf{Cyc} & \textbf{mAP}
& \textbf{Car} & \textbf{Ped} & \textbf{Cyc} & \textbf{mAP}
& \textbf{Car} & \textbf{Ped} & \textbf{Cyc} & \textbf{mAP} \\
\midrule

\multirow{2}{*}{BEVFusion}
& w/o VBS$^2$M 
& 16.52 & 14.37 & 35.18 & 22.02
& 19.85 & 17.42 & 39.26 & 25.51
& 22.03 & 19.34 & 42.15 & 27.84
& 24.16 & 21.08 & 44.37 & 29.87 \\
& w/ VBS$^2$M  
& \textbf{22.36} & \textbf{19.21} & \textbf{40.57} & \textbf{27.38}
& \textbf{24.18} & \textbf{21.06} & \textbf{43.51} & \textbf{29.58}
& \textbf{25.73} & \textbf{22.54} & \textbf{45.32} & \textbf{31.20}
& \textbf{26.84} & \textbf{23.63} & \textbf{47.05} & \textbf{32.51} \\

\midrule

\multirow{2}{*}{RaCFormer}
& w/o VBS$^2$M 
& 34.28 & 15.70 & 25.16 & 25.05
& 37.24 & 18.06 & 29.17 & 28.16
& 39.18 & 19.84 & 32.06 & 30.36
& 41.06 & 21.31 & 34.25 & 32.21 \\
& w/ VBS$^2$M  
& \textbf{39.04} & \textbf{18.71} & \textbf{31.36} & \textbf{29.70}
& \textbf{41.36} & \textbf{20.42} & \textbf{34.08} & \textbf{31.95}
& \textbf{43.25} & \textbf{22.03} & \textbf{36.17} & \textbf{33.82}
& \textbf{44.91} & \textbf{23.34} & \textbf{38.06} & \textbf{35.44} \\

\midrule

\multirow{2}{*}{RCBEVDet}
& w/o VBS$^2$M 
& 29.84 & 13.92 & 24.36 & 22.71
& 32.47 & 16.24 & 28.12 & 25.61
& 34.35 & 18.05 & 30.83 & 27.74
& 36.18 & 19.46 & 33.02 & 29.55 \\
& w/ VBS$^2$M  
& \textbf{34.08} & \textbf{17.42} & \textbf{29.86} & \textbf{27.12}
& \textbf{36.12} & \textbf{19.05} & \textbf{32.41} & \textbf{29.19}
& \textbf{37.85} & \textbf{20.47} & \textbf{34.26} & \textbf{30.86}
& \textbf{39.21} & \textbf{21.76} & \textbf{36.05} & \textbf{32.34} \\

\bottomrule
\end{tabular}}
\end{table*}

\begin{table*}[t]
\centering
\caption{Few-shot target-domain generalization on TJ4DRadSet $\rightarrow$ VoD in terms of AP/mAP (\%). 
The model is trained on the full TJ4DRadSet source dataset and additionally uses different proportions of labeled VoD target-domain training data.}
\label{tab:fewshot_tj_to_vod}
\resizebox{\textwidth}{!}{
\renewcommand{\arraystretch}{1.05}
\begin{tabular}{ll|cccc|cccc|cccc|cccc}
\toprule
\multirow{2}{*}{\textbf{Model}} 
& \multirow{2}{*}{\textbf{Setting}}
& \multicolumn{4}{c|}{\textbf{0\% Target}}
& \multicolumn{4}{c|}{\textbf{10\% Target}}
& \multicolumn{4}{c|}{\textbf{20\% Target}}
& \multicolumn{4}{c}{\textbf{30\% Target}} \\
\cmidrule(lr){3-6} \cmidrule(lr){7-10} \cmidrule(lr){11-14} \cmidrule(lr){15-18}
& 
& \textbf{Car} & \textbf{Ped} & \textbf{Cyc} & \textbf{mAP}
& \textbf{Car} & \textbf{Ped} & \textbf{Cyc} & \textbf{mAP}
& \textbf{Car} & \textbf{Ped} & \textbf{Cyc} & \textbf{mAP}
& \textbf{Car} & \textbf{Ped} & \textbf{Cyc} & \textbf{mAP} \\
\midrule

\multirow{2}{*}{BEVFusion}
& w/o VBS$^2$M 
& 23.63 & 28.15 & 44.36 & 32.05
& 26.54 & 30.76 & 48.05 & 35.12
& 28.71 & 32.88 & 51.26 & 37.62
& 30.84 & 34.71 & 53.95 & 39.83 \\
& w/ VBS$^2$M  
& \textbf{30.12} & \textbf{31.86} & \textbf{51.04} & \textbf{37.67}
& \textbf{32.04} & \textbf{33.57} & \textbf{53.26} & \textbf{39.62}
& \textbf{33.46} & \textbf{35.02} & \textbf{55.13} & \textbf{41.20}
& \textbf{34.82} & \textbf{36.11} & \textbf{56.47} & \textbf{42.47} \\

\midrule

\multirow{2}{*}{RaCFormer}
& w/o VBS$^2$M 
& 29.31 & 31.15 & 37.48 & 32.65
& 32.64 & 33.58 & 40.92 & 35.71
& 34.86 & 35.42 & 43.73 & 38.00
& 36.91 & 37.18 & 46.25 & 40.11 \\
& w/ VBS$^2$M  
& \textbf{36.08} & \textbf{35.02} & \textbf{43.47} & \textbf{38.19}
& \textbf{38.23} & \textbf{36.84} & \textbf{45.72} & \textbf{40.26}
& \textbf{39.75} & \textbf{38.21} & \textbf{47.58} & \textbf{41.85}
& \textbf{41.12} & \textbf{39.67} & \textbf{49.03} & \textbf{43.27} \\

\midrule

\multirow{2}{*}{RCBEVDet}
& w/o VBS$^2$M 
& 24.56 & 26.84 & 43.27 & 31.56
& 27.72 & 29.38 & 47.06 & 34.72
& 29.83 & 31.44 & 50.12 & 37.13
& 31.96 & 33.18 & 52.67 & 39.27 \\
& w/ VBS$^2$M  
& \textbf{30.14} & \textbf{31.53} & \textbf{49.72} & \textbf{37.13}
& \textbf{32.06} & \textbf{33.14} & \textbf{51.83} & \textbf{39.01}
& \textbf{33.54} & \textbf{34.62} & \textbf{53.57} & \textbf{40.58}
& \textbf{34.88} & \textbf{35.91} & \textbf{55.18} & \textbf{41.99} \\

\bottomrule
\end{tabular}}
\end{table*}

\subsection{Few-Shot Target-Domain Results}
To further evaluate the adaptation ability of VBS$^2$M when limited target-domain annotations are available, we conduct few-shot target-domain experiments in both VoD $\rightarrow$ TJ4DRadSet and TJ4DRadSet $\rightarrow$ VoD directions. The model is trained with the full source training set together with 10\%, 20\%, or 30\% labeled samples from the target training set, and is evaluated on the full target test set. This setting examines whether our method can further improve cross-domain adaptation under limited target supervision.

As shown in Table~\ref{tab:fewshot_vod_to_tj}, in the VoD $\rightarrow$ TJ4DRadSet direction, the target-domain performance of all models gradually improves as the target-domain annotation ratio increases from 0\% to 30\%, indicating that limited target data can effectively complement the scenario coverage of source-domain training. With VBS$^2$M, all three detectors achieve higher performance under all target-domain ratios. For example, under the 30\% target setting, BEVFusion, RaCFormer, and RCBEVDet achieve mAP values of 32.51\%, 35.44\%, and 32.34\%, respectively, clearly outperforming their counterparts without VBS$^2$M.

Table~\ref{tab:fewshot_tj_to_vod} reports the few-shot target-domain results in the TJ4DRadSet $\rightarrow$ VoD direction. VBS$^2$M also maintains advantages in the reverse transfer setting. Under the 10\%, 20\%, and 30\% target-domain settings, all three detectors obtain steady improvements after introducing VBS$^2$M. For instance, under the 30\% target setting, the mAP values of BEVFusion, RaCFormer, and RCBEVDet increase to 42.47\%, 43.27\%, and 41.99\%, respectively. These bidirectional results show that VBS$^2$M is effective not only for pure source-only generalization, but also for cross-dataset adaptation with limited target-domain annotations. This is because limited target annotations provide partial real target-domain distribution information, while VBS$^2$M further supplements richer latent scene-shift constraints through source-domain shift mining, improving the model's ability to exploit target-domain scenario variations.

\subsection{Ablation Study}
To understand the contribution of each design in VBS$^2$M, we conduct three groups of ablation studies. 
All ablations are performed using RaCFormer as the base detector, and we report OOD mAP under the bidirectional cross-dataset transfer settings. 
Specifically, we evaluate the cumulative contribution of each component, compare spectral prototype mining with common image and frequency-domain perturbations, and examine whether prototype-guided BEV shifts provide stronger feature-level constraints than unstructured BEV perturbations.

\paragraph{Main Ablations}
To verify the effectiveness of each component, we use RaCFormer as the base detector and evaluate the OOD mAP in both VoD $\rightarrow$ TJ4DRadSet and TJ4DRadSet $\rightarrow$ VoD directions. As shown in Table~\ref{tab:ablation_vbs2m}, the random spectral mask brings only limited improvement, indicating that simple frequency-domain perturbation is insufficient to model cross-scenario variations. Introducing image spectral scene prototypes leads to clear performance gains, suggesting that latent scene patterns mined from source-domain frequency statistics are beneficial for cross-domain generalization. Further incorporating BEV scene shift prototypes improves the results consistently, demonstrating the necessity of modeling the propagation from visual shifts to BEV feature shifts. The full VBS$^2$M achieves the best OOD mAP of 29.70\% and 38.19\% in the two transfer directions, respectively, validating the effects of spectral scene mining, BEV shift modeling, and prototype-guided regularization.

\begin{figure}[t]
    \centering
    \includegraphics[width=\linewidth]{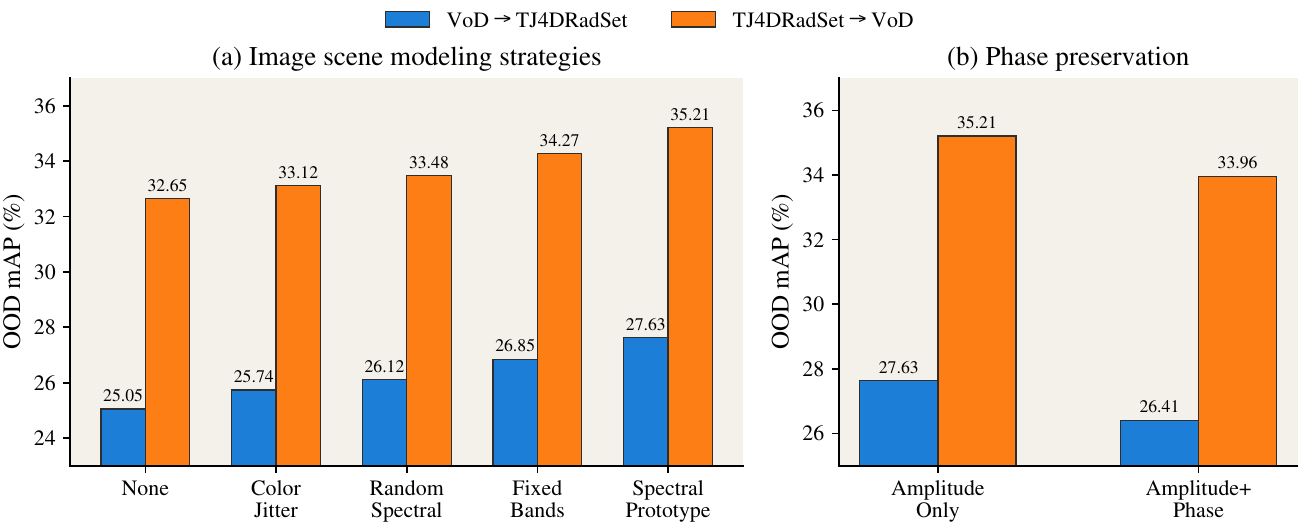}
    \caption{Ablation on image spectral scene modeling. (a) Comparison of image scene modeling. (b) Comparison of spectral reconstruction strategies. }
    \label{fig:image_spectral_ablation}
\end{figure}

\paragraph{Ablation on Image Spectral Scene Modeling}
We further analyze the design of the image spectral scene modeling module. 
As shown in Figure~\ref{fig:image_spectral_ablation}(a), color jitter brings only limited gains, indicating that pixel-space appearance perturbation is insufficient to capture cross-dataset scene gaps. 
Random spectral masking performs slightly better, showing that frequency-domain variation is more relevant to cross-scene generalization. 
However, fixed frequency-band perturbation still relies on manually predefined frequency ranges and cannot adapt to the source-domain scene distribution. 
In contrast, spectral prototype mining achieves the best OOD mAP in both transfer directions, demonstrating the benefit of data-driven spectral scene mining. 
Figure~\ref{fig:image_spectral_ablation}(b) further shows that amplitude-only modulation outperforms perturbing both amplitude and phase, suggesting that preserving phase helps maintain spatial geometry during image scene modulation.

\paragraph{Ablation on BEV Scene Regularization}
We compare different BEV feature regularization strategies on top of spectral prototype mining.
As shown in Table~\ref{tab:bev_regularization_ablation}, random Gaussian noise and channel-wise shift bring only marginal improvements, indicating that unstructured feature perturbations are insufficient to model cross-scenario BEV shifts.
Statistic perturbation performs better by modifying feature distributions, but it still ignores how image-level scene changes propagate into BEV space.
In contrast, prototype-guided BEV shift achieves the best performance in both transfer directions, demonstrating that the mined BEV scene shift prototypes provide more structured and effective regularization for cross-dataset generalization.

\begin{table}[t]
\centering
\caption{Main Ablation of VBS$^2$M on RaCFormer.}
\label{tab:ablation_vbs2m}
\resizebox{\linewidth}{!}{
\renewcommand{\arraystretch}{1.15}
\begin{tabular}{l|ccc|cc}
\toprule
\textbf{Setting} 
& \textbf{Spectral} 
& \textbf{BEV} 
& \textbf{BEV} 
& \textbf{VoD $\rightarrow$} 
& \textbf{TJ4DRadSet $\rightarrow$} \\
& \textbf{Prototype} 
& \textbf{Prototype} 
& \textbf{Reg.} 
& \textbf{TJ4DRadSet} 
& \textbf{VoD} \\
\midrule
Baseline 
&  &  &  
& 25.05 & 32.65 \\

Random Spectral Mask 
&  &  &  
& 26.12 & 33.48 \\

Spectral Prototype Only 
& $\checkmark$ &  &  
& 27.63 & 35.21 \\

Spectral + BEV Prototype 
& $\checkmark$ & $\checkmark$ &  
& 28.52 & 36.74 \\

Full VBS$^2$M 
& $\checkmark$ & $\checkmark$ & $\checkmark$ 
& \textbf{29.70} & \textbf{38.19} \\
\bottomrule
\end{tabular}}
\end{table}

\begin{table}[t]
\centering
\caption{Ablation on BEV scene regularization strategies using RaCFormer. All settings are built on spectral prototype mining. Metric: OOD mAP (\%).}
\label{tab:bev_regularization_ablation}
\resizebox{\linewidth}{!}{
\renewcommand{\arraystretch}{1.12}
\begin{tabular}{l|cc}
\toprule
\textbf{BEV Regularization Strategy}
& \textbf{VoD $\rightarrow$ TJ4DRadSet}
& \textbf{TJ4DRadSet $\rightarrow$ VoD} \\
\midrule
None 
& 27.63 & 35.21 \\

Gaussian Noise 
& 27.91 & 35.48 \\

Channel-wise Shift 
& 28.14 & 35.86 \\

Statistic Perturbation 
& 28.37 & 36.22 \\

Prototype-guided BEV Shift 
& \textbf{29.70} & \textbf{38.19} \\
\bottomrule
\end{tabular}}
\end{table}

\subsection{Sensitivity Analysis}

We analyze the sensitivity of VBS$^2$M to four key hyperparameters: the number of image spectral prototypes $K$, the number of BEV scene shift prototypes $L$, the image spectral modulation strength $\lambda_I$, and the BEV regularization strength $\lambda_B$.
All experiments use RaCFormer as the base detector and report OOD mAP under the two transfer directions, VoD $\rightarrow$ TJ4DRadSet and TJ4DRadSet $\rightarrow$ VoD.
For prototype numbers, we vary $K,L\in\{2,4,8,16\}$.
For shift strengths, we vary $\lambda_I\in\{0.05,0.10,0.20,0.30\}$ and $\lambda_B\in\{0.01,0.03,0.05,0.10,0.20\}$, while keeping other hyperparameters fixed.

\begin{figure}[t]
    \centering
    \includegraphics[width=0.99\linewidth]{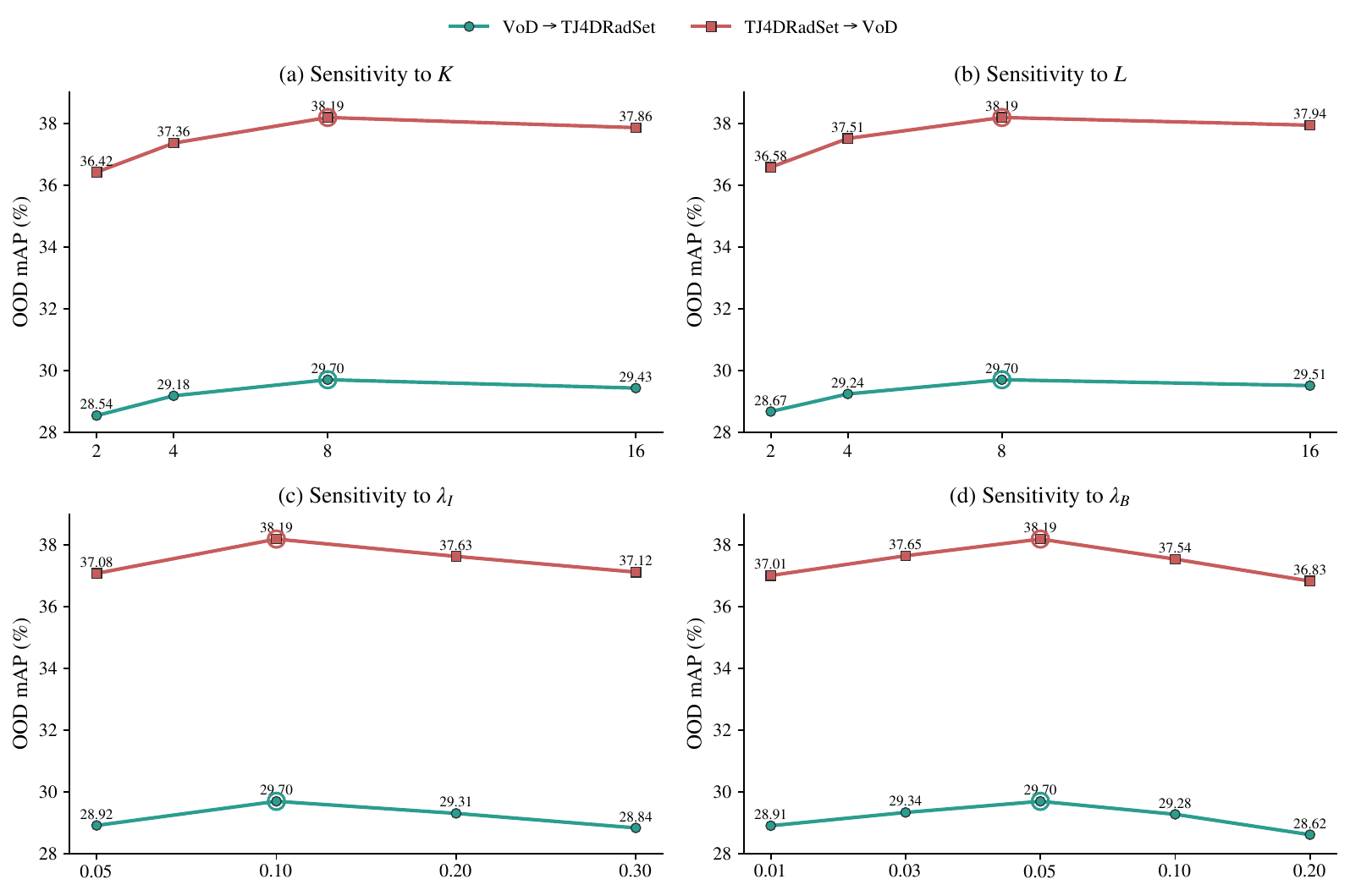}
    \caption{
    Sensitivity analysis of VBS$^2$M.
    (a) Sensitivity to the image spectral prototype number $K$.
    (b) Sensitivity to the BEV scene shift prototype number $L$.
    (c) Sensitivity to the image spectral modulation strength $\lambda_I$.
    (d) Sensitivity to the BEV regularization strength $\lambda_B$.
    }
    \label{fig:sensitivity_analysis}
\end{figure}

As shown in Figure~\ref{fig:sensitivity_analysis}, increasing $K$ and $L$ from 2 to 8 consistently improves OOD mAP in both transfer directions, indicating that multiple prototypes help capture diverse source-domain scene shift patterns.
However, further increasing the prototype number to 16 brings no additional gains and slightly degrades performance, likely because excessive prototypes introduce redundant or fragmented shift modes.
For $\lambda_I$ and $\lambda_B$, moderate values achieve the best results: overly small values provide insufficient scene variation or BEV regularization, while overly large values may distort image statistics or BEV representations.
Based on these results, we set $K=8$, $L=8$, $\lambda_I=0.10$, and $\lambda_B=0.05$ as the default configuration.
\section{Additional Analysis}
\subsection{Visual-to-BEV Shift Magnitude Analysis}
To examine whether VBS$^2$M stabilizes the feature propagation from visual observations to fused BEV representations, we measure how much the BEV feature changes when the input image is replaced by its scene-shifted view. 
Specifically, for each sample, we extract the BEV feature \(f_B\) from the original radar-camera input and the BEV feature \(\hat{f}_B\) from the corresponding scene-shifted image with the same radar input. 
We then compute the visual-to-BEV shift magnitude as \(m_B = 1-\mathrm{Cos}(f_B,\hat{f}_B)\), where a larger value indicates that image-level scene variation causes stronger fluctuations. 

\begin{figure}[t]
    \centering
    \includegraphics[width=\linewidth]{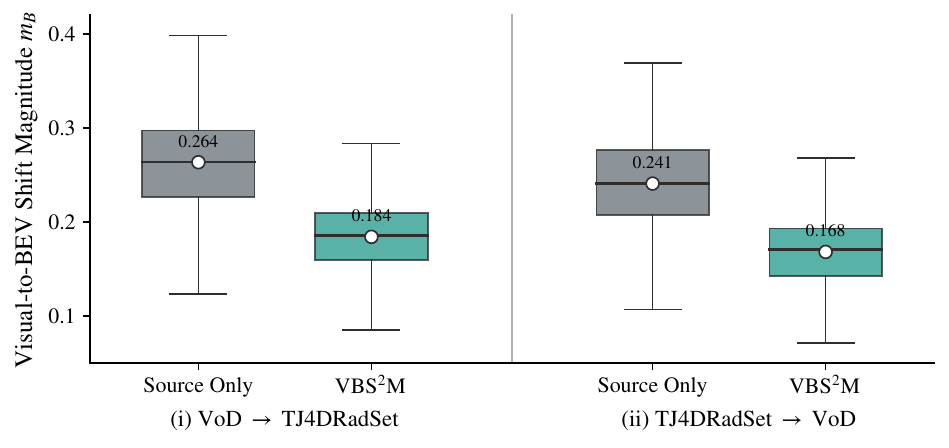}
    \caption{
    Visual-to-BEV shift magnitude analysis.
    Lower values indicate more stable BEV representations under visual scene shifts.
    }
    \label{fig:visual_to_bev_shift_magnitude}
\end{figure}

As shown in Fig.~\ref{fig:visual_to_bev_shift_magnitude}, the Source Only model exhibits larger and more dispersed BEV shift magnitudes in both transfer directions, suggesting that its fused BEV features are highly sensitive to visual scene changes and may amplify image-level distribution gaps during BEV fusion. 
In contrast, VBS$^2$M clearly reduces the shift magnitude and produces a more concentrated distribution. 
This indicates that the proposed spectral scene mining and prototype-guided BEV regularization help constrain how visual scene variations propagate into BEV space, making the fused representation less sensitive to source-specific visual statistics. 
They provide mechanistic evidence that VBS$^2$M does not merely improve detection performance, but also learns more stable visual-to-BEV feature transformations for cross-dataset generalization.

\subsection{Visualization of Spectral Scene Prototypes}

To examine whether VBS$^2$M learns meaningful image-level scene patterns, we visualize the spectral modulation masks generated by the learned image spectral prototypes. 
Specifically, for each spectral prototype \(c_k^I\), we obtain its corresponding frequency modulation mask through \(M_k^I=\mathrm{Reshape}(W_Ic_k^I)\). 
We visualize these masks in the shifted frequency plane, where the center corresponds to low-frequency components and the outer regions correspond to high-frequency components. 
Different spatial regions in the frequency plane therefore reflect different types of image scene statistics, such as illumination, contrast, texture, edge details, and directional frequency patterns.

\begin{figure}[t]
    \centering
    \includegraphics[width=\linewidth]{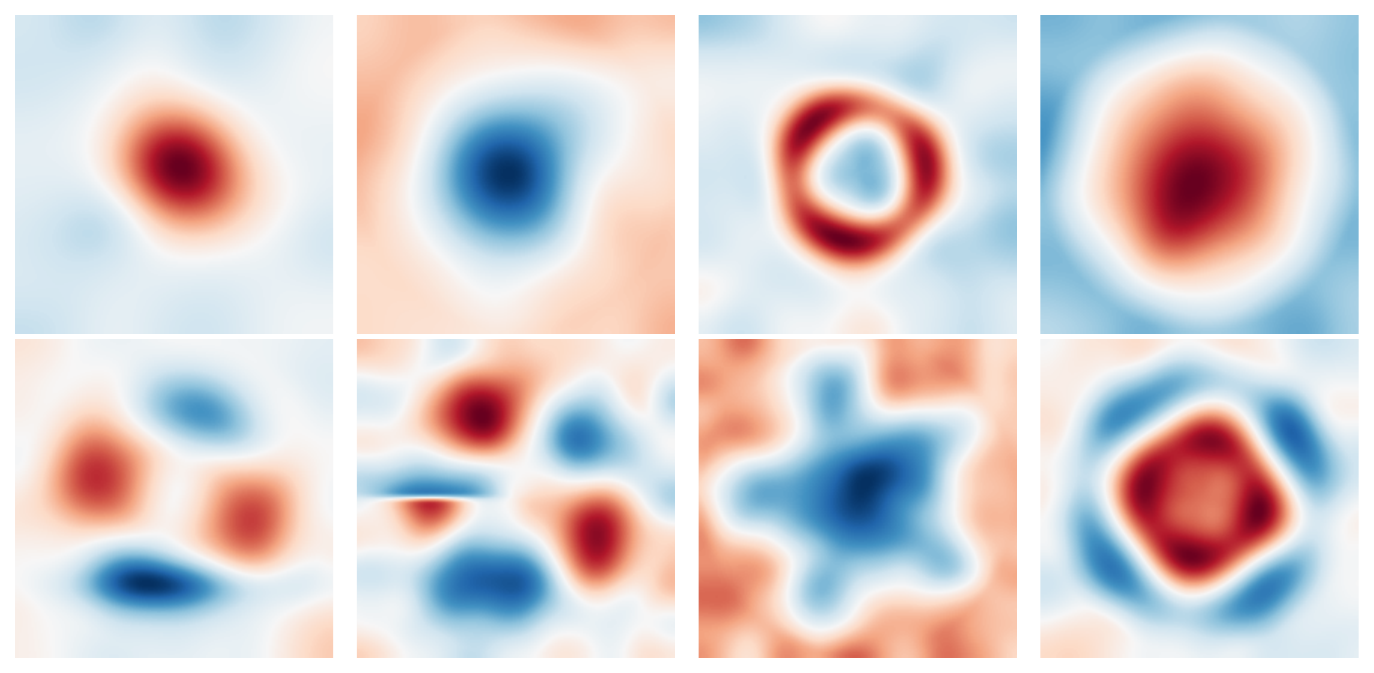}
    \caption{
    Visualization of learned spectral scene prototypes. Each prototype is shown as a frequency modulation mask, with low-frequency components at the center and high-frequency components in the outer regions.
    }
    \label{fig:spectral_prototypes}
\end{figure}

As shown in Fig.~\ref{fig:spectral_prototypes}, the learned prototypes exhibit diverse and structured frequency responses rather than random noise patterns. 
Some prototypes mainly modulate low-frequency regions, indicating that they capture global appearance variations such as illumination and exposure style. 
Some prototypes show clear mid-frequency or high-frequency responses, corresponding to texture complexity, local contrast, edge details, and blur-related changes. 
Other prototypes contain directional frequency structures, suggesting that the model also captures orientation-dependent scene patterns. 

\begin{figure}[t]
    \centering
    \includegraphics[width=\linewidth]{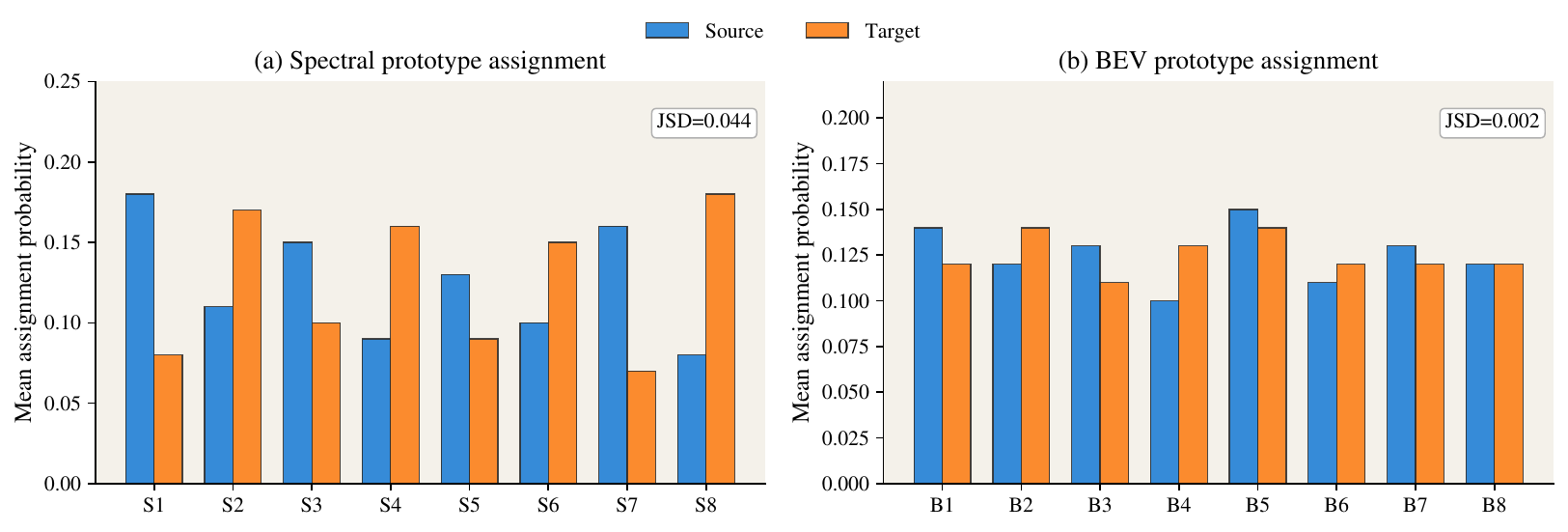}
    \caption{
    Prototype assignment analysis. (a) Mean assignment distributions over image spectral prototypes. (b) Mean assignment distributions over BEV scene shift prototypes.
    }
    \label{fig:prototype_assignment_analysis}
\end{figure}

\subsection{Prototype assignment analysis}
We analyze the average assignment distributions of source-domain and target-domain samples over the image spectral prototypes and BEV scene shift prototypes, in order to verify whether the learned prototypes capture meaningful cross-scenario shift patterns. For each sample, we compute its soft assignment weights over the spectral prototypes and BEV prototypes, and then average them over source and target samples separately. As shown in Fig.~\ref{fig:prototype_assignment_analysis}, source and target samples exhibit clear differences in the image spectral prototype assignments, indicating that the spectral prototypes can reflect visual statistical discrepancies. In contrast, the assignment distributions become closer in the BEV prototype space, suggesting that VBS$^2$M transforms input-level visual scene differences into more stable BEV-level shift patterns.

\subsection{Efficiency Analysis}
\begin{figure}[t]
    \centering
    \includegraphics[width=0.85\linewidth]{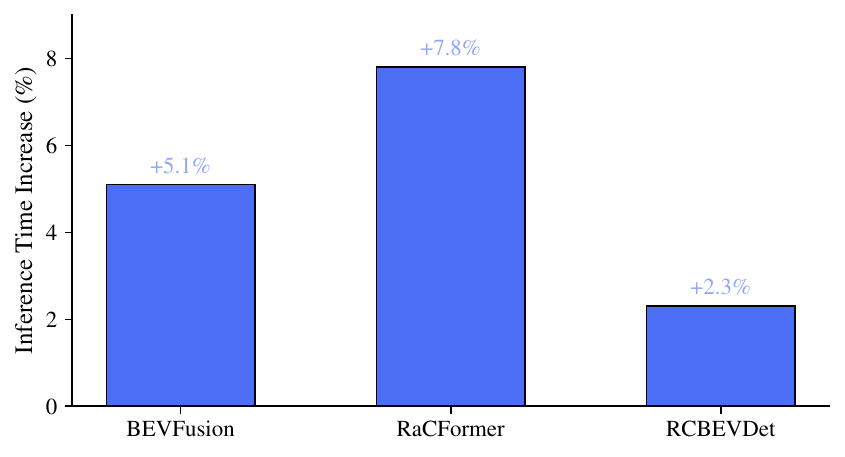}
    \caption{Efficiency analysis of VBS$^2$M. It shows the inference time increase after equipping three representative BEV fusion detectors with VBS$^2$M.}
    \label{fig:efficiency_analysis}
\end{figure}

We evaluate the inference efficiency of VBS$^2$M on three representative BEV fusion detectors. 
As shown in Fig.~\ref{fig:efficiency_analysis}, we report the relative increase in inference time after equipping each detector with VBS$^2$M. 
The results show that the proposed method introduces only minor overhead, with inference time increasing by 5.1\%, 7.8\%, and 2.3\% for BEVFusion, RaCFormer, and RCBEVDet, respectively. 
They indicate that VBS$^2$M improves cross-dataset generalization while maintaining good deployment efficiency.

\section{Conclusion}

\paragraph{Conclusion}
We present VBS$^2$M, a visual-to-BEV sensor shift mining framework for cross-dataset domain generalization in radar-camera BEV detection. Unlike conventional domain generalization methods that rely on generic input augmentation or feature-level regularization, VBS$^2$M models how scene-level sensor shifts propagate from image observations to fused BEV representations. Specifically, VBS$^2$M mines spectral scene prototypes from source-domain image frequency statistics, generates scene-shifted visual views, discovers the corresponding BEV scene shift prototypes, and applies prototype-guided BEV regularization during training. Experiments on bidirectional transfer between VoD and TJ4DRadSet show that VBS$^2$M consistently improves OOD detection performance across multiple radar-camera BEV detectors. Additional ablations and analyses further demonstrate that the gains come not merely from stronger augmentation, but from structured sensor-to-feature shift mining and more stable visual-to-BEV feature propagation.

\paragraph{Future Work}
Future work will explore two directions. First, VBS$^2$M mainly models image-induced scene shifts and their propagation into fused BEV representations; more radar-aware shift mining can be studied without relying on hand-crafted radar perturbations. Second, richer scene context, temporal cues, and uncertainty estimation can be incorporated to discover more fine-grained cross-scenario shift patterns.

\paragraph{Limitation}
This work has several limitations. VBS$^2$M relies on source-domain spectral statistics, so its effectiveness may decrease when the source data lack sufficient scene diversity. Moreover, the learned spectral and BEV prototypes provide structured but coarse shift patterns, whose semantic meanings are not directly tied to specific objects or physical factors. Finally, although inference remains unchanged, the method introduces extra training-time computation for scene-shifted image generation and BEV shift mining.
\clearpage
\bibliographystyle{ieeetr}
\bibliography{icdm}


\end{document}